\documentclass[letterpaper, 10 pt, conference]{ieeeconf} 
\IEEEoverridecommandlockouts
\usepackage[T1]{fontenc}
\usepackage{cite}
\usepackage{amsmath,amssymb,amsfonts}
\usepackage{pifont} 
\usepackage{multirow}
\usepackage{algorithmic}
\usepackage{graphicx}
\usepackage{textcomp}
\usepackage{xcolor}
\usepackage{mathtools}
\usepackage{bm}
\usepackage{tabularx}
\usepackage{subcaption}
\usepackage{float}
\usepackage{stfloats}
\usepackage{svg}

\usepackage{siunitx}
\usepackage{booktabs}

\def\BibTeX{{\rm B\kern-.05em{\sc i\kern-.025em b}\kern-.08em
    T\kern-.1667em\lower.7ex\hbox{E}\kern-.125emX}}

\title{\LARGE \bf
Enabling Dynamic Tracking in Vision-Language-Action Models \\ via Time-Discrete and Time-Continuous Velocity Feedforward
}
\author{Johannes Hechtl$^{1}$, Philipp Schmitt$^{1}$, Georg von Wichert$^{1}$, Wolfram Burgard$^{2}$
\thanks{$^{1}$Siemens Foundational Technologies}
\thanks{$^{2}$Department of Computer Science~\& Artificial Intelligence, University of Technology Nuremberg, Germany.}%
}

\begin{document}

\bstctlcite{IEEEexample:BSTcontrol}
\maketitle
\thispagestyle{empty}
\pagestyle{empty}

\begin{abstract}
While vision-language-action (VLA) models have shown great promise for robot manipulation, their deployment on rigid industrial robots remains challenging due to the inherent trade-off between compliance and responsiveness. Standard Behavior Cloning (BC) approaches predict discrete poses at low frequencies, omitting the velocity and acceleration feedforward terms typically used by low-level compliant controllers. This requires to rely on high stiffness for accurate tracking, thereby sacrificing safe contact dynamics. In this paper, we demonstrate the importance of integrating velocity feedforward terms into VLA policies to resolve this trade-off. We propose two methods for extracting velocity targets from VLAs: a time-discrete finite-difference approximation that serves as a highly effective bridge for existing models, and a continuous Cubic B-Spline action space that natively yields $C^2$ continuous trajectories for high-frequency control. Crucially, both approaches are strictly model-agnostic and compatible with any standard action-chunking architecture, requiring modifications only to teleoperation, data processing, and the low-level controller. We fine-tune the $\pi_{0.5}$ model and evaluate both of our approaches on a demanding, contact-rich cube-in-hole task. Our results indicate that incorporating the velocity feedforward term via finite differences significantly improves task execution speed, while the continuous B-Spline approach maintains high overall success rates and provides a foundation for smoother higher-order derivatives without compromising compliance.
\end{abstract}

% \begin{IEEEkeywords}
% Range sensing, Segmentation, Ground
% \end{IEEEkeywords}

\section{Introduction}
In recent years, Vision-Language-Action (VLA) models have demonstrated significant capabilities in generalist robot manipulation~\cite{brohanRT2VisionLanguageActionModels2023,teamOctoOpenSourceGeneralist2024,kimOpenVLAOpenSourceVisionLanguageAction2024,nvidiaGR00TN1Open2025,intelligence$p_05$VisionLanguageActionModel2025}. This progress has been fueled by the availability of diverse datasets, ranging from large-scale cross-embodiment aggregations~\cite{collaborationOpenXEmbodimentRobotic2024,khazatskyDROIDLargeScaleInTheWild2024,walkeBridgeDataV2Dataset2024} and simulation benchmarks~\cite{meesCALVINBenchmarkLanguageConditioned2022,liEvaluatingRealWorldRobot2024}, to legacy datasets for rigid manipulators~\cite{kalashnikovMTOptContinuousMultiTask2021, dasariRoboNetLargeScaleMultiRobot2020}.

Despite this data diversity, a significant portion of current research utilizes compliant hardware, such as the ViperX arms in the ALOHA setup~\cite{zhaoLearningFineGrainedBimanual2023, zhaoALOHAUnleashedSimple2024}. These manipulators are lightweight, possess inherently weak actuators, and are safe for contact-rich interaction by design. However, their limited payload (approx.~750g) restricts their utility in industrial settings.

\begin{figure}[htbp]
    \centering
    \includegraphics[width=\columnwidth]{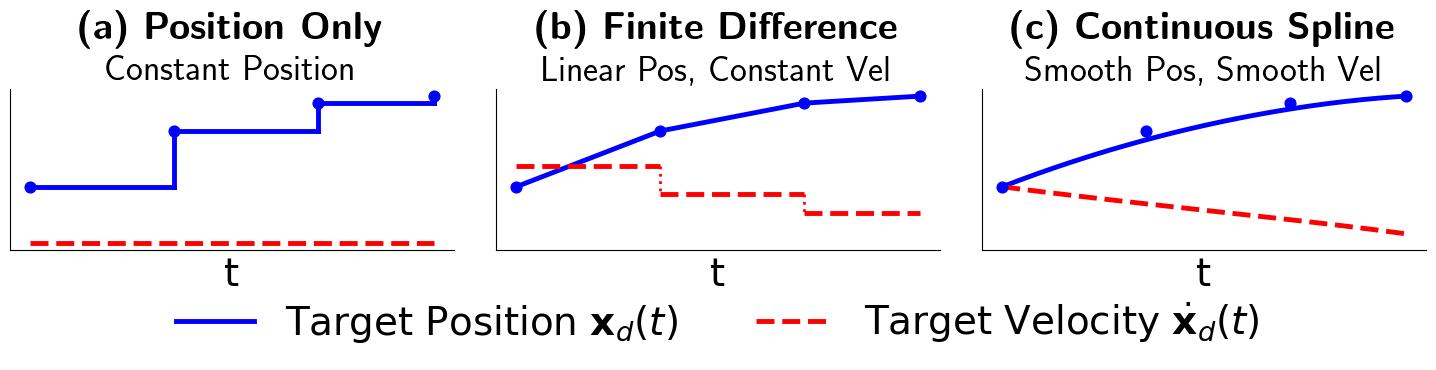}
    \caption{Comparison of target trajectory representations (also called action chunks) for low-level controllers. Blue dots indicate time-discrete policy predictions. 
    (a) Baseline: Standard VLA outputs yield stepwise position references without velocity feedforward. 
    (b) Ours-1: Finite-difference approximation assumes linear interpolation between positions, providing piecewise-constant velocities. 
    (c) Ours-2: Our proposed Cubic B-Spline action space generates $C^2$ continuous position and velocity profiles directly from model predictions.}
    \label{fig:trajectory_comparison}
\end{figure}

Deploying VLAs in industrial environments requires rigid robots with higher payloads and precision, such as the Franka Emika, Kuka LBR, or Universal Robots (UR) series. Performing contact-rich manipulation with these stiff, rigid actuators introduces a risk of damaging the robot or the environment. The standard solution in robotics theory is the implementation of active compliance, specifically impedance or admittance control~\cite{hogan1984ImpedanceControl}. These controllers regulate the dynamic relationship between external forces and robot motion, typically simulating a mass-spring-damper system.

In Behavior Cloning (BC), which forms the basis of VLA models, the fundamental objective is to imitate expert demonstrations. Consequently, good model performance relies on two distinct factors: (1) a high-quality teleoperation system that establishes the upper bound of performance and~(2) a BC formulation capable of accurately capturing and reproducing the demonstrated dynamics.

While compliant controllers are well-established for classical control tasks~\cite{albu-schafferCartesianImpedanceControl2003, haddadinFrankaEmikaRobot2022}, their integration into BC and VLA architectures presents a theoretical challenge. 
Standard BC models typically predict discrete end-effector positions (or other representations like pose deltas or joint positions)~$\mathbf{x}_d$ at a fixed action frequency (e.g., 15--60 Hz)~\cite{collaborationOpenXEmbodimentRobotic2024}. 
However, a second-order impedance controller requires not just a target position, but rather a full reference trajectory including velocity $\dot{\mathbf{x}}_d$ and acceleration $\ddot{\mathbf{x}}_d$ feedforward terms.

The absence of these feedforward terms creates a fundamental trade-off. Without accurate $\dot{\mathbf{x}}_d$ and $\ddot{\mathbf{x}}_d$, the controller relies entirely on the stiffness term proportional to the position error $\mathbf{e} = \mathbf{x}_d - \mathbf{x}$ for tracking. Consequently, to minimize tracking error (lag), one must increase stiffness, thereby reducing compliance. Conversely, to maintain compliance, one must accept tracking errors. This limits the robot's ability to perform tasks that require both high throughput and delicate interaction.

Existing implementations of impedance controllers for VLAs~\cite{proCRISPCompliantROS22025,luoSERLSoftwareSuite2025} predominantly employ zeroth-order control laws or set $\dot{\mathbf{x}}_d = 0$, $\ddot{\mathbf{x}}=0$. Recent methods such as CompliantVLA-adaptor~\cite{zhangCompliantVLAadaptorVLMGuidedVariable2026} and FILIC~\cite{geFILICDualLoopForceGuided2025} attempt to mitigate interaction risks by dynamically modulating stiffness gains based on task context or force feedback. However, while these approaches determine when to be compliant, they do not address how to maintain tracking fidelity during compliant motion. By relying exclusively on position feedback and omitting feedforward velocity and acceleration terms, these methods still accept an inherent tracking lag. While this latency is often tolerated in research settings since it rarely impacts success rates, it imposes a limit on execution speed that restricts industrial deployment, where cost-effectiveness directly correlates with system throughput.

Current software infrastructure typically abstracts away low-level dynamics. For instance, the Deoxys framework for Franka Emika~\cite{zhuVIOLAImitationLearning2023} and standard implementations in Polymetis~\cite{linPolymetis2021}  or the Robot Control Stack(RCS) \cite{juelg2025robotcontrolstacklean} operate primarily as setpoint regulators. To ensure safety within this paradigm, methods like SERL~\cite{luoSERLSoftwareSuite2025} implement reference limiting to saturate velocity during large step changes, prioritizing safe operation over dynamic tracking fidelity.

Alternative approaches, such as the Adaptive Compliance Policy (ACP)~\cite{houAdaptiveCompliancePolicy2024}, learn to modulate stiffness, effectively training the policy to exploit the controller's tracking errors to exert force. However, this learned compliance introduces significant complexity and requires a special teleoperation setup.

Recent works such as BEAST~\cite{zhou2025beast} and FAST~\cite{pertsch2025fastefficientactiontokenization} have also explored B-spline formulations for VLA action spaces. However, these approaches utilize splines primarily as a compression method and tokenizer for discrete action spaces.

In this work, we directly address the tracking-compliance trade-off. We demonstrate that actively providing velocity feedforward terms to the low-level controller can significantly improve VLA performance. To the best of our knowledge, we are the first to integrate an impedance/admittance controller with explicit velocity feedforward into a VLA formulation.

Our contributions are as follows:
\begin{enumerate}
    \item We demonstrate that incorporating a velocity feedforward term in the low-level controller can allow VLA policies to achieve faster execution times without sacrificing necessary compliance.
    \item We propose a discrete velocity approximation method that computes velocity via finite differences between consecutive policy predictions, serving as an effective and easily deployable solution for existing discrete VLA architectures (see Fig.~\ref{fig:trajectory_comparison}).
    \item We introduce a continuous Cubic B-Spline Action Space, predicting spline control points to guarantee $C^2$ continuous trajectories, thereby allowing analytical evaluation of position and velocity at the native high-frequency control rate.
    \item We empirically evaluate the impact of adding velocity feedforward to the robot controller both during teleoperation and rollouts of learned policies.
\end{enumerate}

\section{Methodology}
\label{sec:methodology}

We frame the robot manipulation task as a standard policy learning problem. 
At time step $t$, the policy $\pi$ receives an observation $\mathbf{o}_t$ (e.~g., comprising visual data and the proprioceptive state $\mathbf{q}_t$) and predicts an action $\mathbf{a}_t$. In contemporary Behavior Cloning (BC) approaches utilizing action chunking, $\mathbf{a}_t$  corresponds to a sequence of target poses over a future time horizon at a fixed frequency. To provide the necessary feedforward terms for the low-level compliant controller, we evaluate two distinct, additional representations for the action $\mathbf{a}_t$: a finite-difference approach that operates on these discrete poses, and our proposed B-Spline method, which parameterizes $\mathbf{a}_t$ as the control points of a continuous-time trajectory. Note: neither of these trajectory representations are by themselves novel, but their application to BC is.

\subsection{Cartesian Compliance and Feedforward Dynamics}
\label{subsec:compliance_dynamics}
To enable safe, contact-rich manipulation with rigid industrial robots, we assume active compliance control.  
In the remainder of this work we will use Cartesian compliance control as the running example (while our methodology applies equally to joint-level compliance as well).
Whether implemented as Impedance Control on torque-controlled robots or Admittance Control on position-controlled robots, the desired closed-loop behavior simulates a mass-spring-damper system:
\begin{equation}
    \label{eq:impedance_main}
    \mathbf{\Lambda}_d \ddot{\mathbf{e}} + \mathbf{D}_d \dot{\mathbf{e}} + \mathbf{K}_d \mathbf{e} = \mathbf{F}_{ext}
\end{equation}
where $\mathbf{e}(t) = \mathbf{x}_d(t) - \mathbf{x}(t)$ represents the tracking error between the desired reference $\mathbf{x}_d$ and the actual end-effector position $\mathbf{x}$. $\mathbf{\Lambda}_d$, $\mathbf{D}_d$, and $\mathbf{K}_d$ denote the positive semi-definite desired inertia, damping, and stiffness matrices, respectively, and $\mathbf{F}_{ext}$ represents external forces acting on the robot.
 
In the remainder of this work we will further specify our running example to use an admittance controller, which commands the robot's Cartesian acceleration $\ddot{\mathbf{x}}$ by substituting $\ddot{\mathbf{e}} = \ddot{\mathbf{x}}_d - \ddot{\mathbf{x}}$ into Equation (\ref{eq:impedance_main}) and solving for $\ddot{\mathbf{x}}$:
\begin{equation}
    \label{eq:admittance_cmd}
    \ddot{\mathbf{x}}_{cmd} = \ddot{\mathbf{x}}_d - \mathbf{\Lambda}_d^{-1} \left( \mathbf{D}_d \dot{\mathbf{e}} + \mathbf{K}_d \mathbf{e} - \mathbf{F}_{ext} \right)
\end{equation}

Standard VLA architectures are designed to output only position targets (i.~e.,~$\dot{\mathbf{x}}_d = \ddot{\mathbf{x}}_d = \mathbf{0}$).
At this point we encourage the reader to pause and review the implications of this for Eq.~\ref{eq:impedance_main} and Eq.~\ref{eq:admittance_cmd}:
Even for free space motions where there is no interaction with the environment ($\mathbf{F}_{ext} = 0$) it is not possible to have both low tracking errors (low~$\mathbf{e}$, requires high gains~$\mathbf{K}_d$) and compliance (low gains~$ \mathbf{K}_d$) at the same time.

This motivates us to include the velocity feedforward terms~$\dot{\mathbf{x}}_d$ both during data collection via teleoperation and as part of the trajectory representation of policies learned from that data.
As illustrated in Fig.~\ref{fig:trajectory_comparison}, we define two methods to extract velocity targets from a VLA and compare them against the standard baseline:

\begin{enumerate}
    \item \textbf{Position-Only Targets (Baseline):} The policy predicts discrete position targets $\mathbf{x}_d$, while the velocity feedforward term is strictly set to zero ($\dot{\mathbf{x}}_d = \mathbf{0}$). As shown in Fig.~\ref{fig:trajectory_comparison}a, this yields stepwise position references, forcing the low-level controller to rely entirely on the stiffness term and positional error for tracking.
    \item \textbf{Time-Discrete Velocity Approximation (Method 1):} The policy predicts time-discrete positions, and velocity is approximated during inference by calculating the finite difference between consecutive positions: $\dot{\mathbf{x}}_t \approx (\mathbf{x}_t - \mathbf{x}_{t-1}) / \Delta t$. As depicted in Fig.~\ref{fig:trajectory_comparison}b, this approach effectively bridges existing position-only VLA outputs with velocity-aware controllers, provided the model is trained on data recorded with the same velocity-aware controller. However, it results in piecewise-constant velocity profiles that can introduce minor discontinuities.
    \item \textbf{Time-Continuous Action Space (Method 2):} The actions in the dataset are transformed into Cubic B-Spline control points such that the action space intrinsically models continuous trajectories. As illustrated in Fig.~\ref{fig:trajectory_comparison}c, the policy predicts analytical curves, allowing smooth position and velocity profiles to be sampled directly at the high control frequency $f_{ctrl}$.
\end{enumerate}

\subsection{Method 2: Cubic B-Spline Action Space}
\label{subsec:bsplines}
To provide the necessary smooth, higher-order derivatives for Method~2, we propose an action space based on Cubic B-Splines. Instead of predicting robot positions directly, the VLA model predicts a sequence of control points $\mathbf{P} = \{\mathbf{P}_0, \dots, \mathbf{P}_n\}$, which define a time-continuous reference trajectory $\mathbf{x}_d(t)$.

A B-spline curve of degree $k$ is defined by its control points $\mathbf{P}_i$ and a knot vector $\mathbf{t} = \{t_0, \dots, t_{n+k+1}\}$. The position on the curve at any time $t \in [t_k, t_n]$ is given by the linear combination of basis functions:
\begin{equation}
    \mathbf{x}_d(t) = \sum_{i=0}^{n} B_{i,k}(t) \mathbf{P}_i,
\end{equation}
where $B_{i,k}(t)$ are the B-spline basis functions of degree $k$. These functions are defined recursively \cite{deboorPracticalGuideSplines1978}:

\begin{align}
B_{i, 0}(t) =  &
\begin{cases}
    1, & \textrm{if $t_i \le t < t_{i+1}$}\\
    0, & \textrm{otherwise}
\end{cases}\\
B_{i, k}(t) = &\frac{t - t_i}{t_{i+k} - t_i} B_{i, k-1}(t)
                 \nonumber \\ & + \frac{t_{i+k+1} - t}{t_{i+k+1} - t_{i+1}} B_{i+1, k-1}(t)
\end{align}

For our implementation, we use cubic splines ($k=3$), which ensure $C^2$ continuity (continuity of position, velocity, and acceleration). This allows the low-level controller to analytically derive the feedforward velocity $\dot{\mathbf{x}}_d(t)$ and acceleration $\ddot{\mathbf{x}}_d(t)$ at any arbitrary time $t$ by differentiating the basis functions.
\begin{align}
    \dot{\mathbf{x}}_d(t) & = \sum_{i=0}^{n} \dot{B}_{i,3}(t) \mathbf{P}_i\\ \ddot{\mathbf{x}}_d(t) & = \sum_{i=0}^{n} \ddot{B}_{i,3}(t) \mathbf{P}_i
\end{align}

During inference at step $\tau$, the VLA predicts a local set of control points $\mathbf{P}_\tau$ defining the trajectory for the upcoming horizon. The low-level robot controller queries this spline to obtain $(\mathbf{x}_d, \dot{\mathbf{x}}_d)$ at the native $f_{ctrl}$ rate. This explicitly decouples the low-frequency action generation rate $f_{action}$ from the high-frequency control domain.

\begin{figure}[htbp]
    \centering
    \includegraphics[width=1.05\columnwidth]{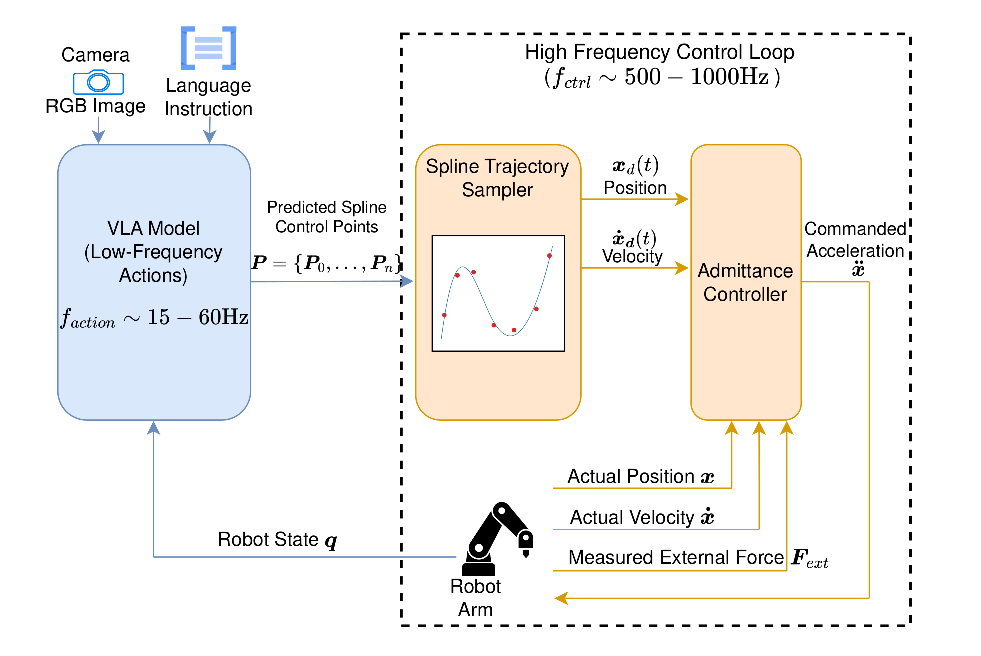}
    \caption{Workflow illustrating Cubic B-Spline inference. The VLA outputs control points, which are continuously sampled by the high-frequency controller.}
    \label{fig:spline inference}
\end{figure}

To train the B-Spline model, raw teleoperation data $\mathcal{T}_{raw}$, recorded at high frequency, must be converted into B-Spline control points. We extract these by solving a Least Squares optimization over the entire recorded trajectory. This offline processing yields a standard behavior cloning dataset $\mathcal{D}_{spline} = \{(\mathbf{o}_t, \mathbf{P}_t)\}_{t=1}^N$, allowing the use of off-the-shelf imitation learning algorithms. Furthermore, because cubic B-splines have local support, any point on the curve depends only on four adjacent control points in both directions.

\section{Results}
To evaluate the efficacy of integrating explicit velocity feedforward terms into VLA-driven manipulation, we structure our experiments into two main analyses: evaluating the impact on human teleoperation performance, and assessing the resulting policy execution.

\subsection{Experiment Setup}
\begin{figure}[htbp]
    \centering
    \includegraphics[width=\columnwidth]{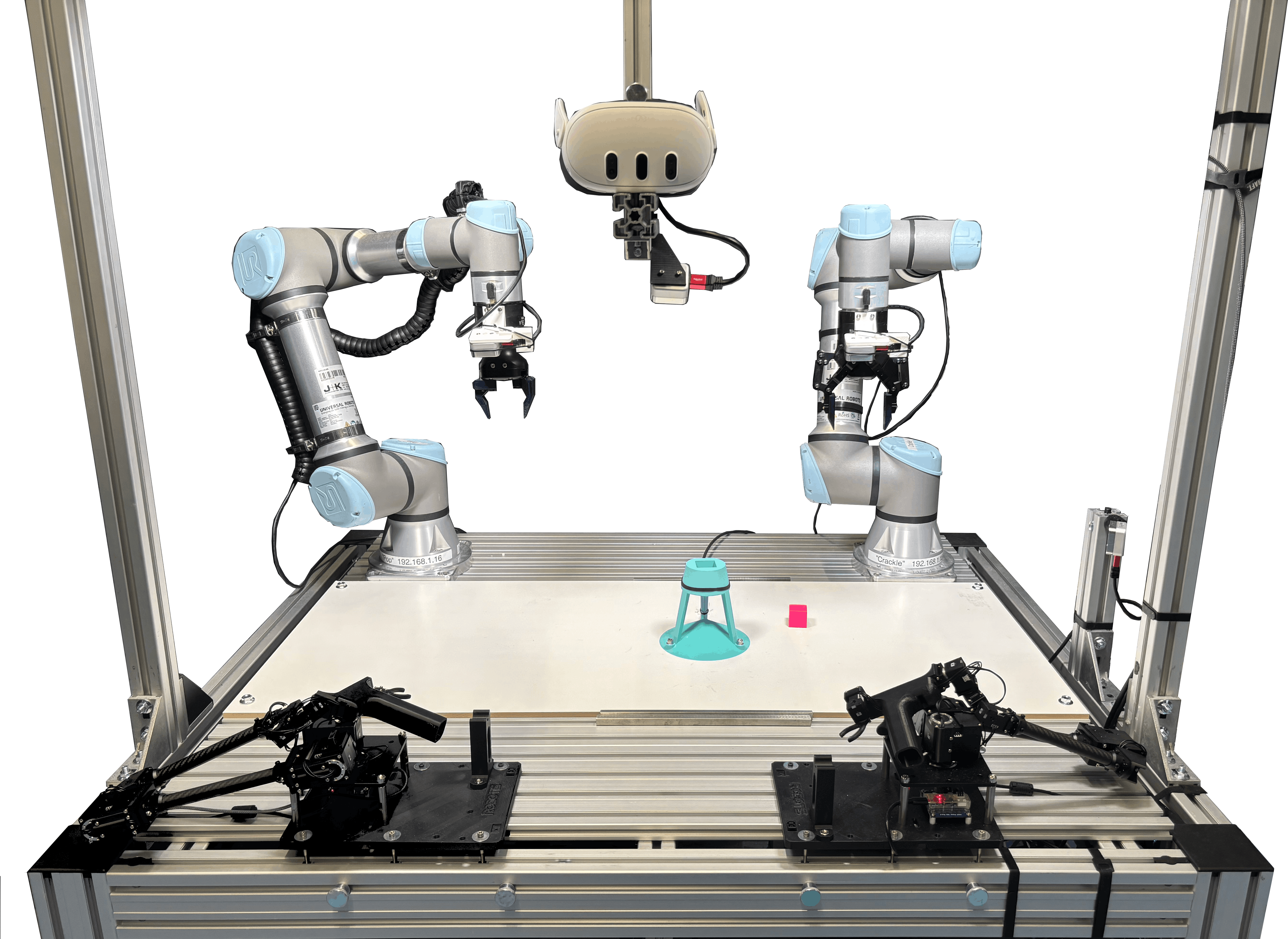}
    \caption{Experimental setup. Teleoperation is performed using two OMY-L100 devices.}
    \label{fig:teleop_setup}
\end{figure}

\begin{figure}[htbp]
    \centering
    \includegraphics[width=\columnwidth]{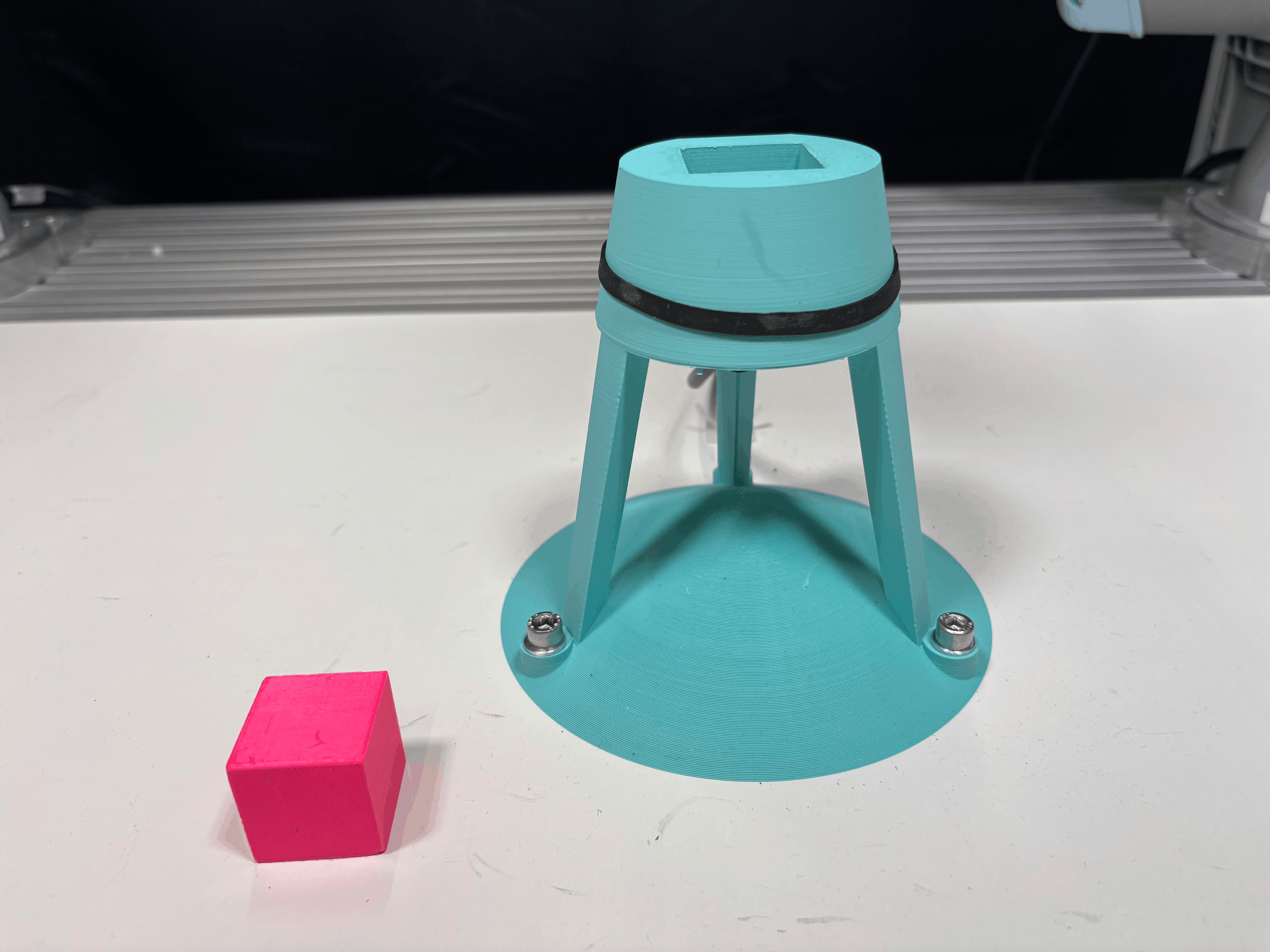}
    \caption{Closeup of the cube-in-hole task. A strict $1\unit{mm}$ tolerance necessitates compliance.}
    \label{fig:task_closeup}
\end{figure}

We evaluate our methods on a contact-rich cube-in-hole task, as shown in Figures~\ref{fig:teleop_setup} and \ref{fig:task_closeup}. The environment consists of a plastic fixture with a square opening. The objective is to pick up a cube and insert it into the hole. Because there is only a $1\unit{mm}$ margin between the cube and the fixture, the insertion phase requires compliant behavior to avoid both jamming and destroying the fixture. 

Upon successful insertion, the cube falls onto a cone beneath the fixture, which deflects it to a random resting position within the workspace. This necessitates a large, transfer motion for the subsequent trial. Consequently, high task throughput in this environment requires two control requirements: (1) fast and accurate trajectory tracking during the transfer phase, and (2) highly compliant behavior during the insertion phase.

Our teleoperation setup uses two OMY-L100 devices.
These teleoperation devices provide desired position values~$\mathbf{x}_d$ at a high frequency (>500Hz) which allows to compute the desired velocities~$\dot{\mathbf{x}}_d$ using low pass filtered finite differences.
While incorporating an acceleration feedforward term $\ddot{\mathbf{x}}_d$ could theoretically further improve tracking fidelity, obtaining reliable acceleration references from a teleoperation setup is practically challenging due to noise amplification during double differentiation. 
Consequently, to prevent high-frequency noise from destabilizing the controller, our present implementation leverages $\dot{\mathbf{x}}_d$ but assumes~$\ddot{\mathbf{x}}_d = \mathbf{0}$. 

We use two Universal Robots (UR) as examples for industrially used, rigid robots. 
The robots are controlled via the \texttt{ur\_rtde}~\cite{lindvigUr_rtdeInterfaceControlling2025} library and we implement an active, Cartesian admittance controller as in Eq.~\ref{eq:admittance_cmd}.

To demonstrate the compatibility of our methodology with state-of-the-art architectures, we fine-tune the $\pi_{0.5}$ VLA model~\cite{intelligence$p_05$VisionLanguageActionModel2025} for the policy evaluations. 
All models are trained for $40,000$ steps with a batch size of $32$.

Standard inference delays often produce jumps at action chunk boundaries. These discontinuous trajectories become highly problematic when attempting dynamic robot motions. To mitigate this during policy execution, we employ Real Time Chunking (RTC)~\cite{blackRealTimeExecutionAction}, which leverages the inherent inpainting capabilities of flow-matching policies to enable smooth transitions between action chunks.

\subsection{Teleoperation Results}
\begin{figure}[htbp]
    \centering
    \includegraphics[width=\columnwidth]{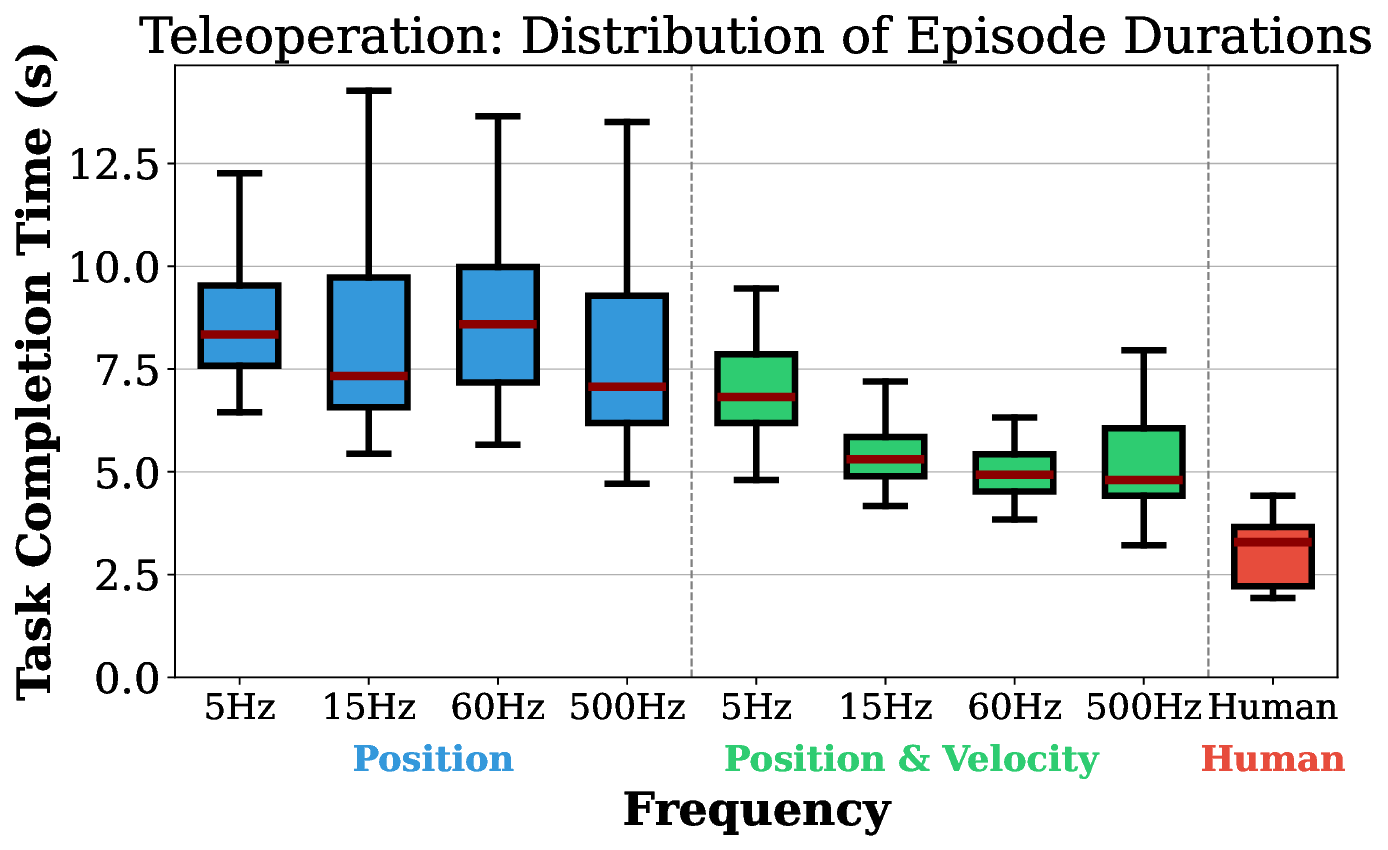}
    \caption{Distribution of episode durations for teleoperation across different controller configurations. We include human performance for comparison.}
    \label{fig:teleop_success_length_distribution}
\end{figure}

To evaluate our first hypothesis—that integrating a velocity feedforward term enables faster execution and more accurate tracking—we analyzed the speed of the teleoperated demonstrations. A human operator executed the task for $10$-minute intervals under various controller configurations and action frequencies. For this experiment, the priority was fast task execution, and no measures for creating a structured demonstration dataset were taken. Figure~\ref{fig:teleop_success_length_distribution} illustrates the distribution of task completion times.

For the baseline (position-only), teleoperation performance remained largely consistent regardless of the reference update rate. Even when artificially bottlenecked to a highly delayed $5\unit{Hz}$ update rate, the operator adapted to the system lag, and the low-level controller sufficiently smoothed the commanded trajectory to maintain task success.

Conversely, the velocity-aware controller proved more reactive. As detailed in Table~\ref{tab:ttest_results_teleop}, t-tests confirm that the velocity feedforward approach yields significantly lower task completion times compared to the position-only baseline across all evaluated frequencies ($p < 0.001$). Task completion times for the velocity-aware configurations exhibited less variance, as the operator could execute large transfer motions quickly and accurately without fighting the controller's inherent tracking lag.

However, this increased reactivity meant that the system was more sensitive to extreme delays. Specifically, we observed a noticeable performance drop (an increase in mean completion time) when the action rate was restricted to $5\unit{Hz}$ compared to higher frequencies. Ultimately, the consistent speed advantage when using a velocity-aware controller establishes a higher upper bound for the data quality provided to the VLA model.

\begin{table*}[htbp]
    \centering
    \begin{tabular}{lcccc}
        \hline
        Frequency & Position (Baseline) & Position \& Velocity (Method) & $t$-statistic & $p$-value \\
        \hline
        $5\unit{Hz}$   & $9.58 \pm 3.97$ ($n=68$) & $7.64 \pm 3.00$ ($n=88$)  & $-3.45$  & $<0.001$ \\
        $15\unit{Hz}$  & $8.48 \pm 3.04$ ($n=83$) & $5.85 \pm 2.27$ ($n=120$) & $-7.02$  & $<0.001$ \\
        $60\unit{Hz}$  & $9.31 \pm 3.19$ ($n=84$) & $5.34 \pm 1.41$ ($n=122$) & $-12.06$ & $<0.001$ \\
        $500\unit{Hz}$ & $8.37 \pm 3.49$ ($n=76$) & $5.37 \pm 1.55$ ($n=121$) & $-8.20$  & $<0.001$ \\
        \hline
    \end{tabular}
    \caption{Summary statistics and one-tailed t-test results of Task Completion Times (Seconds) for Position-Only and Velocity-Aware Controllers during Teleoperation. The Velocity Aware Controller leads to significantly shorter Task Execution Times for all tested action frequencies.}
    \label{tab:ttest_results_teleop}
\end{table*}

\subsection{Policy Task Execution Speed}
\begin{figure}[thpb]
    \centering
    \includegraphics[width=\columnwidth]{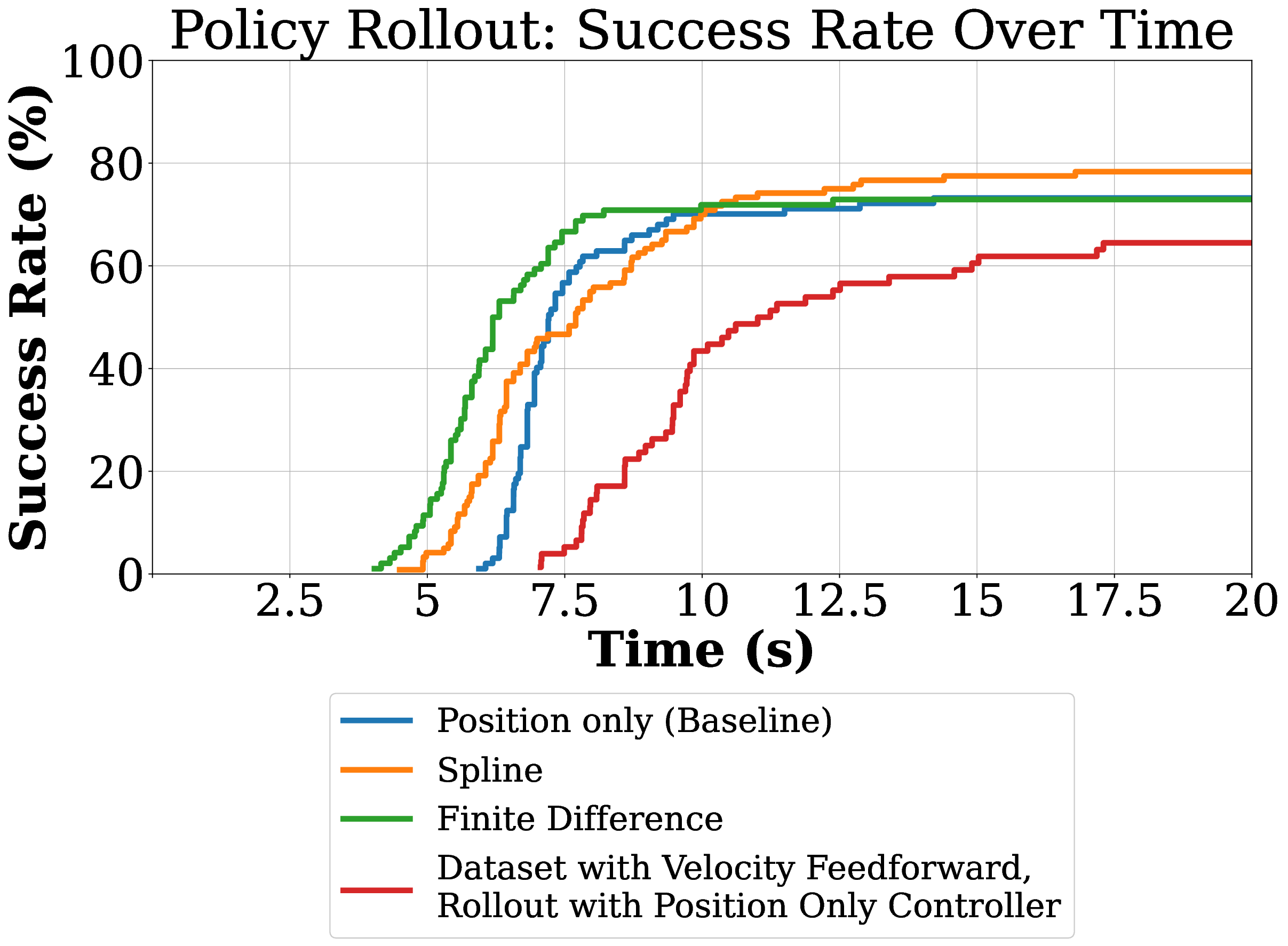}
    \caption{Cumulative success rate over task execution time for different configurations}
    \label{fig:success_rates_over_time}
\end{figure}

To analyze the impact of trajectory representation on task execution speed, Figure~\ref{fig:success_rates_over_time} presents the cumulative success rate over time, and Table~\ref{tab:execution_time_stats} summarizes the corresponding statistical analysis.

The Finite Difference method (Time-Discrete Velocity Approximation) yields the fastest task execution ($M=6.05$\,s) and significantly outperforms the baseline ($M=7.35$\,s) with $p < 0.001$. Successful trials for this method begin around $4$ seconds and plateau near its final success rate by approximately $8$ to $10$ seconds.

The Cubic B-Spline approach, while demonstrating some capability for rapid successful rollouts before the $6$-second mark, exhibits higher variance and a slightly higher mean execution time ($7.45$ seconds) compared to the baseline ($7.35$ seconds). Consequently, it does not yield a statistically significant improvement in execution speed. Despite this slower and less consistent temporal performance, the Spline policy ultimately achieves a high overall success rate of $79.20\%$ by the end of the $20$-second cutoff.

\begin{table*}[htpb]
    \centering
    \begin{tabular}{lcccccc}
        \hline
        Method & Mean (s) & Var. & $N$ & $t$-stat & $p$-value$^*$ & Sig. ($\alpha=0.05$) \\
        \hline
        Baseline & $7.35$ & $1.99$ & $71$ & -- & -- & -- \\
        Finite Difference & $6.05$ & $1.71$ & $70$ & $-5.70$ & $<0.001$ & Yes \\
        Cubic B-Spline & $7.45$ & $4.75$ & $94$ & $0.32$ & $0.626$ & No \\
        Ablation (Velocity-Feedforward Dataset, \\Position Only Rollout) & $10.01$ & $6.10$ & $49$ & $7.47$ & $<0.001$ & Yes \\
        \hline
        \multicolumn{7}{p{0.7\textwidth}}{\footnotesize $^*$ One-tailed $p$-value testing for a decrease in execution time compared to the baseline. For the ablation we instead test for an increase in execution time.}
    \end{tabular}
    \caption{Task Execution Time Statistics and T-Test Results vs. Baseline}
    \label{tab:execution_time_stats}
\end{table*}

We attribute this inconsistent temporal performance to the interaction between the spline formulation and Real-Time Chunking (RTC). While RTC mitigates jumps between action chunks, it does not completely eliminate them. For the Spline method, these residual jumps can lead to momentarily large, sampled velocity values, producing rapid, jerky motions in the robot. We hypothesize that these jerky motions introduce an unintended benefit during the insertion phase: if the manipulator fails to perfectly align the cube initially, these rapid motions (wiggling) may help overcome the $1\unit{mm}$ tolerance and find the hole, thus improving the final success rate.

Finally, we demonstrate the necessity of matching the inference controller to the teleoperation setup. 
We evaluated an ablated policy that was trained on data collected with a velocity-aware controller, but executed during rollout using a position-only controller. 
As expected, this mismatch severely degrades performance. 
This configuration yields the slowest mean execution time of $10.01$ seconds---with successes only starting after $7.5$ seconds---and achieves the lowest overall success rate among the evaluated models.
This ablation highlights the importance of matching the interpolation strategy during rollouts of policies with the availability of velocity feedforward terms (or lack thereof) during data collection.

\section{Conclusion and Future Work}
In this work, we addressed the inherent trade-off between compliant interaction and dynamic tracking in VLA-driven robotic manipulation. We demonstrated that integrating explicit velocity feedforward terms into the low-level compliance controller enables rigid robots to execute fast transfer motions without sacrificing the compliance necessary for contact-rich tasks. 

We introduced two distinct approaches for bridging discrete VLA predictions with continuous control dynamics: a highly deployable finite-difference approximation, and a continuous Cubic B-Spline action space. Importantly, both methods are entirely model-agnostic. They are fully compatible with any standard action-chunking VLA, requiring changes strictly in data collection, pre/post-processing, and the low-level controller, without altering the underlying neural network architecture. We evaluated both methods on a demanding cube-in-hole task. The finite-difference approach provides a simple, easily deployable method to significantly improve task execution times over position-only baselines.

Future research will focus on integrating full acceleration feedforward ($\ddot{\mathbf{x}}_d$) into the control loop. Given the extreme noise sensitivity of double differentiation, the guaranteed $C^2$ continuity inherently provided by our proposed Cubic B-Spline action space shows promise for enabling stable, high-speed, compliant manipulation.

\section*{Acknowledgment}
We used Gemini (Google, Gemini 3.1 Pro) to aid literature searches and minor grammar/style edits. All content was reviewed and revised by the authors, and the final manuscript is entirely our own work.

\bibliographystyle{IEEEtran}
\bibliography{IEEEabrv,literature}
\end{document}